\documentclass[10pt,journal,letterpaper,compsoc]{IEEEtran}

\usepackage{lineno,graphicx,amsmath,amssymb}
\usepackage{floatflt}
\usepackage{geometry}
\usepackage{algorithm}
\usepackage{algorithmic}

\graphicspath{{figs/}{pics/}{}}
\DeclareGraphicsExtensions{.jpg,.pdf,.mps,.png}

\def\argmin{\mathop{\rm argmin}}
\def\RR{\mathbb R}

\newcommand{\beps}{\boldsymbol{\epsilon}}
\newcommand{\bphi}{\boldsymbol{\varphi}}
\newcommand{\bo}{\boldsymbol{1}}
\newcommand{\bc}{\boldsymbol{c}}

\newcommand{\bp}{\boldsymbol{p}}
\newcommand{\br}{\boldsymbol{r}}

\newcommand{\bx}{{\bf x}}
\newcommand{\by}{{\bf y}}
\newcommand{\bz}{\boldsymbol{z}}
\newcommand{\bu}{{\bf u}}
\newcommand{\bw}{{\bf w}}

\title{Face Detection with a 3D Model}

\author{Adrian Barbu, Nathan Lay, Gary Gramajo\\
\IEEEcompsocitemizethanks{\IEEEcompsocthanksitem A. Barbu, and G. Gramajo are with the Department of Statistics, Florida State University, Tallahassee, Florida 32306, USA, Fax: 850-644-5271, Email: abarbu@stat.fsu.edu, ggramajo@stat.fsu.edu.
\IEEEcompsocthanksitem N. Lay is with the National Institutes of Health, Bethesda, MD 20892, USA. Email: nathan.lay@nih.gov}
}

\begin{document}

\maketitle

\begin{abstract}
This paper presents a part-based face detection approach where the spatial relationship between the face parts is represented by a hidden 3D model with six parameters. The computational complexity of the search in the six dimensional pose space is addressed by proposing meaningful 3D pose candidates by image-based regression from detected face keypoint locations. The 3D pose candidates are evaluated using a parameter sensitive classifier based on difference features relative to the 3D pose. A compatible subset of candidates is then obtained by non-maximal suppression. Experiments on two standard face detection datasets show that the proposed 3D model based approach obtains results comparable to or better than state of the art.
\end{abstract}


\section{Introduction}

In designing a good face detector we have mainly two choices. We might want a simple classifier (e.g. a sliding window classifier) and not care about the inner face representation or the face parts. In this case we would need to train it with features that have good discrimination power and tens of thousands of training examples to be able to cover the large variability of the faces in images due to 3D pose, illumination direction, face occlusions and other factors. For example, the face detector from \cite{zhang2009winner} was trained with 100,000 faces and the detector from \cite{mathias2014face} with more than 26,000 faces and their perturbations.

If we want a more interpretable model where the face parts are taken into consideration, then we are faced with the computational problem of enforcing dependencies between the face parts. A common way to handle this problem is through the DPM framework  \cite{felzenszwalb2010object}, and has been applied to face detection in many recent works such as \cite{mathias2014face,yan2014fastest,zhu2012face}. Another way is by image-based regression, which has been used for face alignment in a number of works \cite{burgos2013robust,cao2012face,dollar2010cascaded,ren2014face} and has been also used for face detection recently in the Joint Cascade\cite{chen2014joint}.

Besides being more interpretable, an added benefit of a part based model is that the face parts (eyes, mouth, nose, ears, chin, etc) have much smaller variability because they have simpler 3D shapes and appearances than the whole face. Thus a face detection system based on these parts could be trained with fewer training examples. For example the DPM model \cite{zhu2012face} has been trained with only 900 faces and obtains very good face detection results. 
The DPM based model uses 18 planar models to represent the part configurations for many possible out of plane face rotations. At test time, it tries all of them and returns the best scoring configurations above a threshold.
This 2D approach leads to the question: what obstacles are there in using a single 3D face model instead of all these 2D models? We argue that the obstacles are mostly computational. The 2D models used in the DPM approaches have a tree structure so that dynamic programming can be applied to obtain a globally optimal configuration. Thus the 2D models make some modeling compromises (many 2D tree-based models instead of a single 3D model) in order to be guaranteed that the global optimum is found. The 2D models used in the face alignment based approaches are not restricted to tree structure and use image-based regression to search for the optimal configuration in the high dimensional shape space.

\begin{floatingfigure}[r]{4.5cm}
\vspace{-2.mm}
\centering
\includegraphics[width=4.5cm]{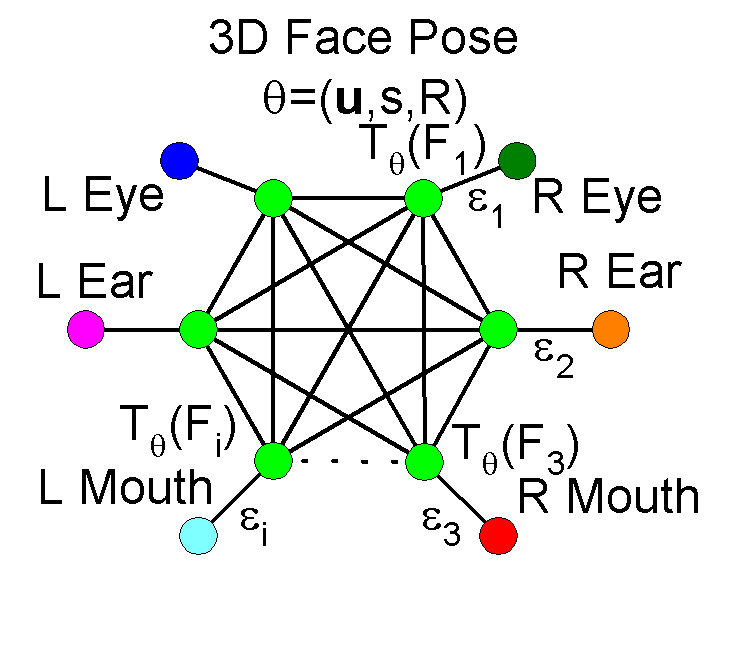}
\vskip -8mm
\caption{The face keypoints are fully connected by a simplex in our 3D model.}
\label{fig:diagMRF}
\vspace{1mm}
\end{floatingfigure}In this paper we investigate an approach that uses a rigid 3D model to represent the interactions between the face parts and image based regression to search for it in images. The 3D model contains a rigid 6 parameter face 3D pose and independent deformations for the face parts from the locations predicted by the 3D pose, as illustrated in Figure \ref{fig:diagMRF}. Many face parts can be missing, thus this model can be seen as an AND/OR graph \cite{wu2011numerical} where the 3D face is the noisy AND of the face parts \cite{zhao2011image}. 
This model is a simplex, not a tree, and dynamic programming cannot be used for exact inference. Instead, the 6D pose space for each face will be searched using data-driven proposals made by image-based regression from the detected face part locations. These proposals are then evaluated by an energy model and the lowest energy configuration is the final result. In order for this approach to work well, the face parts need to be detected accurately. 

\vspace{-1mm}
{\bf Contributions.} 
This paper brings the following contributions:

\noindent -- It presents a face representation that lies in a six-dimensional space, which is larger than the usual 3.1 dimensional space of position, scale, and a number of discrete poses. Such a larger dimensional space poses learning challenges to avoid overfitting and computational challenges in finding the hidden parameters quickly.

\noindent -- It presents a novel computational paradigm that offers a combinatorial number of ways for detecting a face by attempting to detect many face parts and requiring only a few successful detections. This approach avoids relying on a cascade with a single computational path, which could fail if not trained appropriately.  

\noindent -- It introduces a novel set of local selected features that are computed based on the candidate 3D pose and have been selected during training from a larger pool by feature selection. 

\noindent -- It introduces a simple parameter sensitive model with a 1D parameter that is virtually as fast as its parameter insensitive counterpart and can be trained by stochastic gradient descent on millions of observations.

A detection cascade locks-in its losses at each level, since anything rejected at any level cannot be recovered. On the other hand, the noisy AND approach presented in this paper is more robust to detection failures than a cascade, as illustrated in the following example. If we assume that each face keypoint could be detected with probability at least 0.9, then the probability that a face has at most 4 face keypoints detected out of 9 is $\sum_{k=0}^4 { 9 \choose k}\cdot 0.9^k \cdot 0.1^{9-k}<0.001$. Thus if a face is considered detected when it has at least 5 out of 9 keypoints detected, then the probability for detecting a face based on its keypoints is at least $0.999$, by the above computation. So even if the keypoint detections are not very reliable, it is unlikely that many of them will fail at the same time. At the same time, a cascade using one keypoint (e.g. face center) as an intermediate level would detect only 90\% of the faces.

\vspace{-2mm}
\subsection{Related Work}

There are different types of approaches to face or object detection and we will relate to the most relevant ones, even though this list might not be nearly close to complete.

\noindent{\bf Multi-view models.}  Some works approach the problem of detecting 3D objects with multiple viewpoints by having separate models for a range of possible views. In contrast, our work uses a single 3D model that controls the 2D positions of the features that are extracted for the face classifier, and the feature weights are also controlled by the 3D pose in the parameter sensitive face classifier.

 The works \cite{viola2004robust,mathias2014face,yang2014aggregate} use rigid classifiers based on different types of predefined features such as Haar \cite{viola2004robust}, HOG \cite{dalal2005histograms}, ACF \cite{dollar2014fast} and combinations. Very good results were obtained in \cite{mathias2014face} and \cite{li2015convolutional}  by training with  many deformations of the positive examples.

Many works \cite{felzenszwalb2010object,hejrati2012analyzing,zhu2012face,yan2014fastest} use deformable part-based models, where the relationships between parts are organized in a tree for obtaining a computationally tractable models. In this work, the parts are only related to the 3D pose, which connects all of them through a high-order simplex, as illustrated in Figure \ref{fig:diagMRF}.

In \cite{hejrati2012analyzing} a 2D  part configuration is detected using a version of the deformable part model \cite{felzenszwalb2010object} and then a 3D pose and shape is inferred from the 2D configuration. In contrast, our work directly uses the 3D pose to represent the relative positions of the parts without going through an intermediate 2D model.

\noindent{\bf 3D view based models.} Some works \cite{payet2011contours,su2009learning} divide the view sphere into a number of sectors and collect templates for each view. Given a new candidate object, detection is obtained by template matching. Our work is not template based, but is based on a parameter sensitive classifier that uses features extracted relative to a 3D model.

\noindent{\bf 3D models.} Our work resembles \cite{liebelt2010multi, hu2012learning} in that features are extracted based on a 3D model and the object pose hypothesis. However, these approaches use complex inference algorithms (one based on EM and the other based on dynamic programming) while our approach uses regression to propose data-driven candidates from multiple channels. Furthermore, our approach does not need any synthetic 3D models, since it constructs the 3D model from training images by energy minimization. Moreover, none of these works was used for face detection.

\noindent{\bf Cascade approaches.} Some of the most successful face detection approaches \cite{chen2014joint,li2015convolutional} are based on a detection cascade. Since a cascade locks in the losses at each stage, it has an upper limit on the detection rate. For example, the Cascade-CNN cannot reach an 86\% detection rate on FDDB. Our approach is not based on a cascade but has candidates proposed through multiple channels. The combinatorial verification (e.g. requiring four face parts detected out of 9) gives more flexibility and allows many part detectors to fail while still detecting a face successfully. Consequently, our face detector can reach higher detection rate, as high as 89\% or more on FDDB.

\noindent{\bf Face alignment.} Pose candidates have been previously proposed by image based regression in the shape regression machine \cite{zhou2007shape} and for face alignment \cite{burgos2013robust,cao2012face,dollar2010cascaded,ren2014face}, however, they are not based on a 3D model. The Cascade-CNN \cite{li2015convolutional} uses a convolutional neural network (CNN) to improve alignment of the detected face bounding boxes, an approach also not based on a 3D model.

Our work generalizes the Local Binary Features \cite{ren2014face} to local (non-binary) versions that are extracted relative to our 3D model. Moreover, we use a parameter sensitive classifier for scoring the candidates instead of a standard classifier.

The Joint Cascade \cite{chen2014joint} obtains very good face detection results by alternating classification with face alignment by regression, both steps using the LBF features \cite{ren2014face}. Our work differs in many ways. First, we use a 3D model instead of a 2D model. Second, we detect many keypoints in a bottom-up step and use them to propose many 3D pose candidates, instead of starting with a "mean face", this way we can obtain higher detection rates. Third, we use a parameter sensitive classifier instead of a standard classifier, which adapts the feature weights to the 3D pose. It is possible that by using one or more face alignment steps we could further improve the detection rate in the low false positive region.

\noindent{\bf Parameter sensitive classifiers.} Parameter sensitive classifiers were introduced in \cite{yuan2007parameter} for Boosting and linear SVM, and in \cite{yuan2011learning} for SVM with multiplicative kernels. While the multiplicative kernel formulation is generic and can be used with multi-dimensional parameters, it might be too computationally expensive for a face verification classifier. This is why we introduced a simple formulation for the one-dimension parameter representation that can be solved by direct energy minimization.

\noindent{\bf Face detection with pose estimation.} Another notable work is the face detection and pose estimation with energy based models \cite{osadchy2007synergistic} which uses a Convolutional Neural Network to directly map the input image patch into a pose manifold for faces and outside the manifold for non-faces. It would be interesting to see how this work compares to the current state of the art methods on the FDDB and AFW datasets.

\vspace{-2mm}
\section{Parameter Sensitive Model}\label{sec:parmodel}

Parameter sensitive models are models that take a feature vector $\bx \in \RR^M$ and a parameter $\theta\in \RR^p$ to obtain a prediction $s(\bx;\theta)$.

{\bf Parameter sensitive linear model.} For example the dependence on $\bx$ could be linear 
\vspace{-3mm}
\begin{equation}
s(\bx;\theta)=\bw(\theta)^T\bx=\sum_{i=1}^M w_i(\theta)x_i \label{eq:scorelbf}
\vspace{-2mm}
\end{equation} with the linear weights $\bw(\theta)$ depending on the parameter $\theta$. Such a model is more computationally efficient than a  SVM classifier with multiplicative kernels \cite{yuan2011learning} since it avoids kernel multiplication with many support vectors.

In our application the parameter $\theta$ is one dimensional, so we discretize the range of $\theta$ into $B_\theta$ equal intervals $[b_i,b_i+1]$ to represent
 $w_i(\theta)$ as piecewise constant $w(\theta)=w_{ik}$ for $\theta\in [b_k,b_{k+1})$. These parameters are collected in the matrix $W=(\bw_1,...,\bw_M), \bw_i\in \RR^{B_\theta}$.

\begin{figure}[htb]
\vspace{-4mm}
\centering
\includegraphics[height=3.5cm]{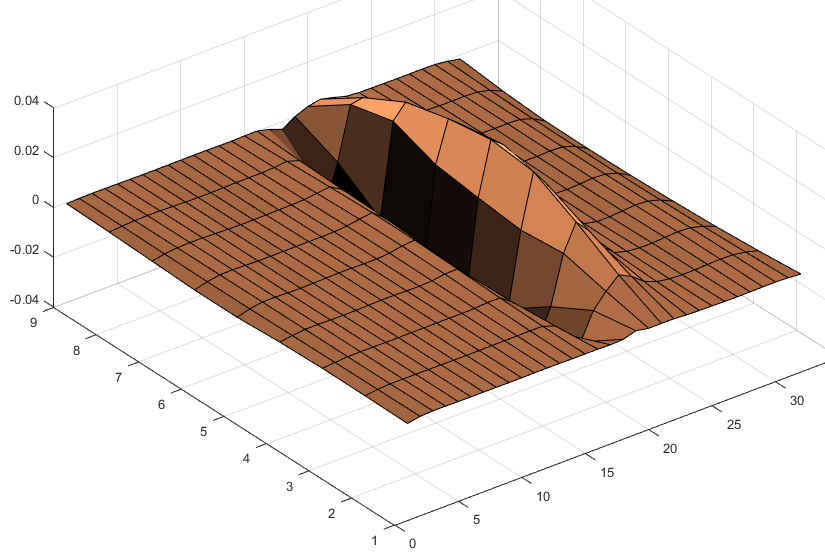}
\vskip -2mm
\caption{One of the bivariate response functions $f_i(x_i,\theta)$ for the nonlinear model \eqref{eq:scorelsf}.}
\label{fig:coeffs2d}
\vspace{-3mm}
\end{figure}

{\bf Non-linear model.} We can introduce nonlinearity without kernels as a sum of bivariate parameter sensitive response functions:
\vspace{-2mm}
\begin{equation}
s(\bx;\theta)=\sum_{i=1}^M f_i(x_i, \theta) \label{eq:scorelsf}
\vspace{-2mm}
\end{equation}
where the function $f_i(x_i, \theta)$ depends on one variable $x_i$ of vector $\bx$ and the parameter $\theta$. An example of a bivariate response function $f_i(x_i,\theta)$ that make up the nonlinear model  \eqref{eq:scorelsf} is shown in Figure \ref{fig:coeffs2d}.

By discretizing the range of $\theta$ into $B_\theta$ bins as above and the range of $x_i$ into $B_\bx$ bins,  we use piecewise constant bivariate response functions $f_i(x_i,\theta)=w_{ijk}$ if $x_i$ is in bin $j$ and $\theta$ is in bin $k$. Again, these parameters are collected in the matrix $W=(\bw_1,...,\bw_M)$ with $\bw_i \in R^{B_{\theta}B_{\bx}}$.

In both cases we write the parameter sensitive model as $s_W(\bx;\theta)$ to emphasize that its parameters form the matrix $W$.

\vspace{-2mm}
 \subsection{Training the Parameter Sensitive Model} \label{sec:optpar}
\vspace{-1mm}
 
The training examples  $(B_i,\bx_i,\theta_i), i=\overline{1,N}$ contain rectangles (boxes) $B_i$, feature vectors $\bx_i$ and the parameters $\theta_i$. The ground truth is given as a set of face boxes $f_j$. Let $o_{ij}$ be the overlap of box $B_i$ with face box $f_j$ if they belong to the same image, otherwise is zero. Let $o_i=\max_j o_{ij}$ be the maximum overlap of box $B_i$ with a face from the same image.

\noindent {\bf Training Cost Function.} Inspired by \cite{wan2015end}, the cost function to be minimized for training:
\vspace{-2mm}
 \begin{equation}
 \begin{split}
E(W)&=\hspace{-1mm}\sum_{j}\hspace{-1mm}\sum_{i,o_{ij}>0.5}\hspace{-4mm}\frac{2o_{ij}-1}{s_j}L(s_W(\bx_i,\theta_i))\\
& + \mu\hspace{-2mm}\sum_{i,o_i<0.3}\hspace{-3mm}L(-s_W(\bx_i,\theta_i)) \hspace{-0.5mm}+\hspace{-1mm}\sum_{i=1}^M \rho(\bw_i)
\end{split}
 \end{equation}
 contains sums of loss terms for each true positive and can be written as a sum of per-image loss terms.
The difference from  \cite{wan2015end} is that the positive boxes (that overlap at least 0.5 with a face) are weighted based on their overlap and then normalized by $s_j=\sum_{i,o_{ij}>0.5}(2o_{ij}-1)$ so that the total weight is 1 for each true face.

The prior $\rho(\bw)$ encourages smooth changes of the parameters between parameter bins
\vspace{-2mm}
\begin{equation}
\rho(\bw)=s\|\bw\|^2+c \sum_{k=2}^{B_\theta-1} (w_{k+1}+w_{k-1}-2w_{k})^2 \label{eq:prior2}
\vspace{-2mm}
\end{equation}
for the linear model and of parameter bins and variable bins for the nonlinear model:
\vspace{-2mm}
\begin{equation}
\begin{split}
\rho(\bw)\hspace{-1mm}=&\hspace{-.5mm}s\|\bw\|^2\hspace{-1.5mm}+\hspace{-1mm}c \hspace{-1mm}\sum_{j=1}^{B_\bx}\hspace{-1.5mm}\sum_{k=2}^{B_\theta-1} \hspace{-1.5mm}(w_{j,k\hspace{-.5mm}+\hspace{-.5mm}1}\hspace{-1mm}+\hspace{-1mm}w_{j,k\hspace{-.5mm}-\hspace{-.5mm}1}\hspace{-1mm}-\hspace{-1.3mm}2w_{jk})^2\\ \label{eq:prior3}
&+\hspace{-1mm}d \hspace{-.1mm}\sum_{k=1}^{B_\theta}\hspace{-1.5mm}\sum_{j=2}^{B_\bx-1} \hspace{-1.5mm}(w_{j\hspace{-.5mm}+\hspace{-.5mm}1,k}\hspace{-1mm}+\hspace{-1mm}w_{j\hspace{-.5mm}-\hspace{-.5mm}1,k}\hspace{-1mm}-\hspace{-1mm}2w_{jk})^2
\end{split}
\end{equation}

\begin{figure}[htb]
\vspace{-6mm}
\centering
\includegraphics[height=3.5cm,width=5cm]{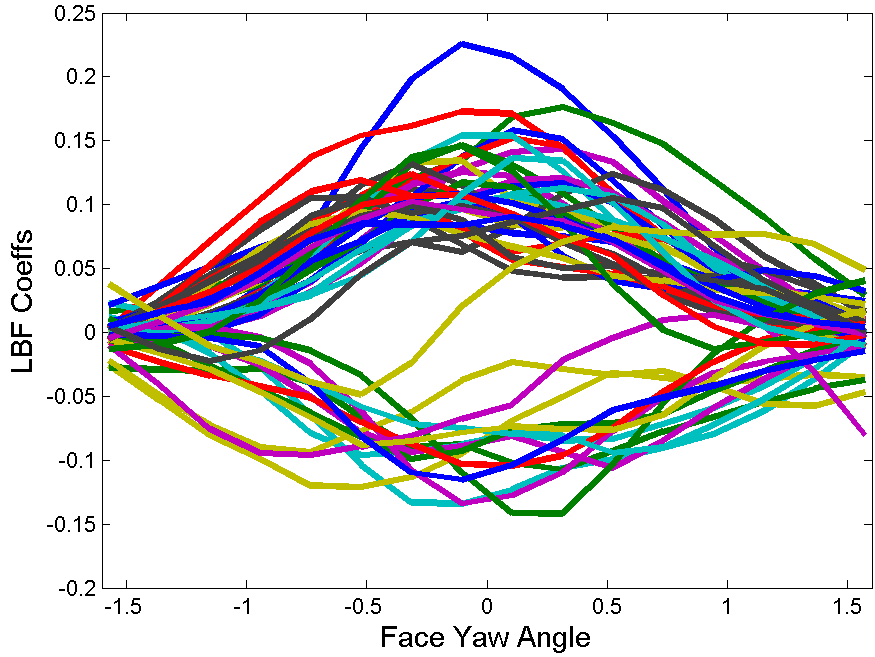}
\vskip -2mm
\caption{Top 50 coefficients by total variation, for a parameter sensitive linear model.}
\label{fig:coeffs}
\vspace{-3mm}
\end{figure}

An example of top learned coefficients for the LBF features is given in Figure \ref{fig:coeffs}, where each curve represents the coefficient of one keypoint across 16 yaw angle bins.

\noindent {\bf Optimization.} Observe that the cost function $E(W)$ is convex when the loss function $L(x)$ is convex (such as the logistic or the hinge loss).

However, we will use the Lorenz loss function \cite{barbu2013feature}
\vspace{-3mm}
\begin{equation}
L(x)=\begin{cases}
\ln(1+(x-1)^2) &\text{ if } x<1\\
0 &\text{ else}
\end{cases} \label{eq:lorenz}
\end{equation} 
which is more robust to outliers than the logistic and hinge losses and was observed to work very well in this application.
 
Optimization is done by Feature Selection with Annealing \cite{barbu2013feature}, a fast iterative procedure that alternates parameter updates with progressive removal of variables according to a schedule. The parameter update step consists of one epoch of stochastic gradient descent with momentum 0.9, where the a batch contains all training examples from one image, thus the parameters are updated one image at a time.

The variable removal schedule is linear, after each parameter update epoch the same number of variables are removed, such that the desired number of variables are left after 20 epochs. In our application we start with 9000 variables and are left with 3000 after 20 epochs.

\vspace{-2mm}
\section{Face Detection Using a 3D Model}\label{sec:3dface}

Given an image, the goal is to find the faces with keypoints and 3D poses. 

\begin{figure}[ht]
\vspace{-4mm}
\centering
\includegraphics[width=5.8cm]{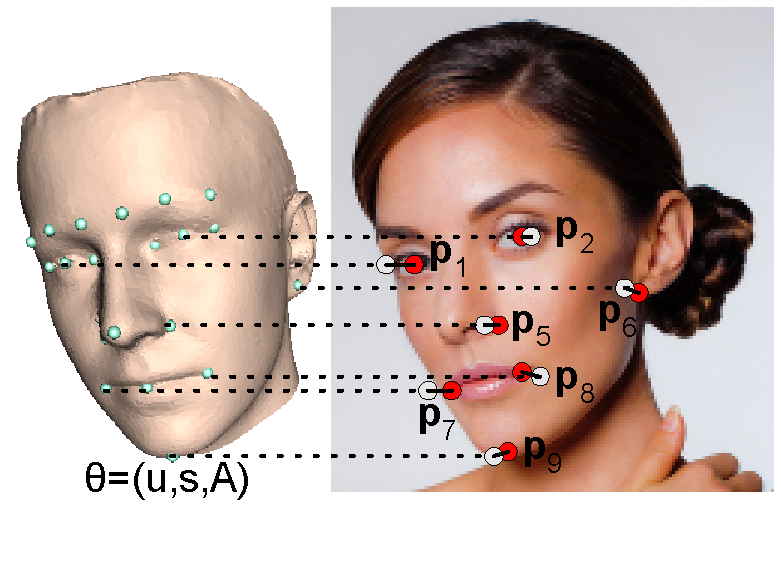}
\vskip -7mm
\caption{A face is represented as a pair of a 3D pose $\theta\hspace{-1mm}=\hspace{-1mm}(\bu,\hspace{-1mm}s,\hspace{-1mm}A)$ and a 2D point configuration $P\hspace{-1mm}=\hspace{-1mm}(\bp_1,...,\bp_L)$.}
\label{fig:3dmodel}
\vspace{-3mm}
\end{figure}
\noindent{\bf  Face Representation.} The face has $L$ keypoints that are 2D points in the image, $P=(\bp_1,...,\bp_L), \bp_i\in \RR^2$. 
The face 3D pose is represented as a projected rigid transformation $T_\theta:\RR^3\to \RR^2$ with parameters $\theta=(\bu,s,A)$ consisting of 2D translation  $\bu\in \RR^2$, scale $s$ and 3D rotation matrix $A$, and defined as $T_\theta(\bx)=\bu+s\pi(A\bx)$, where $\pi:\RR^3\to\RR^2, \pi(x,y,z)=(x,y)$ is the projection on the $(x,y)$ plane.
Thus each face is represented as a pair $F=(P,\theta)$ of the 2D keypoints $P=(\bp_1,...,\bp_L)$ and the 3D pose $\theta=(\bu,s,A)$.
\begin{figure*}[t]
\centering
\includegraphics[width=\linewidth]{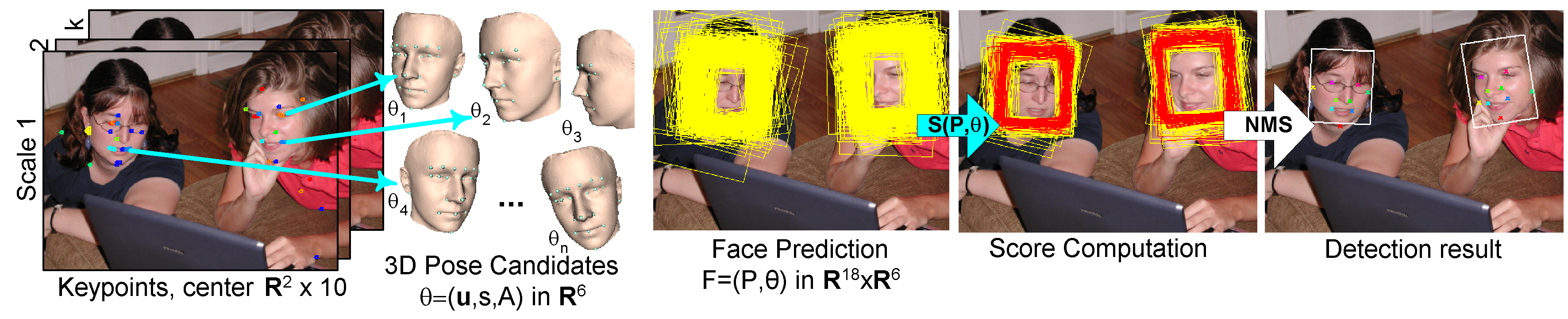}
\vskip -4mm
\caption{Face detection using a 3D model. The face keypoints are detected independently and used to propose 3D pose candidates $\theta=(\bu,s,A)\in \RR^6$. Face candidates are predicted from the rigid model and evaluated using the score $S(P,\theta)$. The detected faces are obtained by non-max suppression.}
\label{fig:diag}
\vspace{-6mm}
\end{figure*}

\noindent{\bf  Face 3D model.} The 3D face model consists of $L$ 3D keypoints in a rigid configuration that can be written as a $3\times L$ matrix $R=(\br_1,...,\br_L), \br_i\in \RR^3$. 
The 2D configuration $P$ of the face keypoints in the image is related to the pose $\theta$ through the relation  
\vspace{-2mm}
\[
\bp_i=T_\theta(\br_i)+\epsilon_i, i=\overline{1,L}
\vspace{-2mm}
\] where $T_\theta$ is the 3D face pose defined above and $\epsilon_i\in\RR^2$ are independent deformations for each keypoint. We will write this in matrix form as:
\vspace{-2mm}
\begin{equation}
P=T_\theta(R)+\beps \label{eq:3dface}
\vspace{-2mm}
\end{equation}

This relation is illustrated in Figure \ref{fig:3dmodel} where the gray dots are the predicted point locations $T_\theta(\br_i)$ and $\bp_i$ are the actual point locations.

\vspace{-2mm}
\subsection{Energy Model}

For any face $F=(P,\theta)$ let $B_F$ be the bounding box of the points $P$. The best configuration of faces is obtained by energy minimization:
\vspace{-2mm}
\[
\begin{split}
(F_1, ..., F_n)&=\argmin_{n, F_1, ..., F_n}(E_{data}(F_1,...,F_n)\\
&+E_{prior}(n,F_1,...,F_n))
\end{split}
\]
The data term 
\vspace{-4mm}
\[
E_{data}(F_1,...,F_n)=\displaystyle{\sum_{j=1}^n }(\tau-S(P_j,\theta_j))
\vspace{-3mm}
\] is based on the scores $S(P_j,\theta_j)$ for the faces $F_j=(P_j,\theta_j)$ in the image
and the parameter $\tau$ that controls the minimum score for a valid detection. The score function $S(P,\theta)$ is defined in more detail in Section \ref{sec:score}.
The prior 
\vspace{-4mm}
\begin{equation}\begin{split}
E_{prior}(n,\hspace{-1mm}F_1,...,\hspace{-1mm}F_n)&=
\hspace{-1mm}\sum_{j=1}^n f(\|P_j-T_{\theta_j}(R)\|) \\ \label{eq:prior}
&+E_{ovr}(n,P_1,...,P_n)
\end{split}\vspace{-3mm}
\end{equation}
has a coherence term between the poses $\theta_j$ and the points $P_j$ of the face $F_j=(P_j,\theta_j)$ and a term $E_{ovr}(n,P_1,...,P_n)$ that 
enforces the constraints that the bounding boxes $B_{F_j}, j=\overline{1,n}$ have small overlap with each other.

\vspace{-2mm}
\subsection{Inference Algorithm}

The inference algorithm is illustrated in Fig. \ref{fig:diag} and consists of the following steps:
\vspace{-1mm}
\begin{enumerate}
\item Bottom-up detection of the face keypoints.
\item Generation of 3D pose candidates $\theta_1,...,\theta_n$ from the keypoints.
\item Generation of the face candidates by pruning the 3D pose candidate $\theta_j$ using the coherence term.
\item Computation of the face scores $S(P_j,\theta_j), j=\overline{1,n}$ and removal of low scoring candidates.
\item Non-maximal suppression to output a set of high score candidates that satisfy the overlap constraints, greedily minimizing the energy $E(n, \theta_1, ..., \theta_n)$.\vspace{-1mm}
\end{enumerate}
These steps are described in the next subsections.

\subsubsection{Detecting face keypoints}\label{sec:kpts}

The $L=9$ face keypoints used in this work are the eye centers, nose sides, mouth corners, bottom ears, and chin. We also detected the center of the face bounding box to handle small faces where the keypoints are not clearly visible. The keypoints are detected on a Gaussian pyramid with 4 scales per octave and minimum image size of $24\times 24$.
\begin{figure}[ht]
\vspace{-3mm}
\centering
\includegraphics[width=7.4cm]{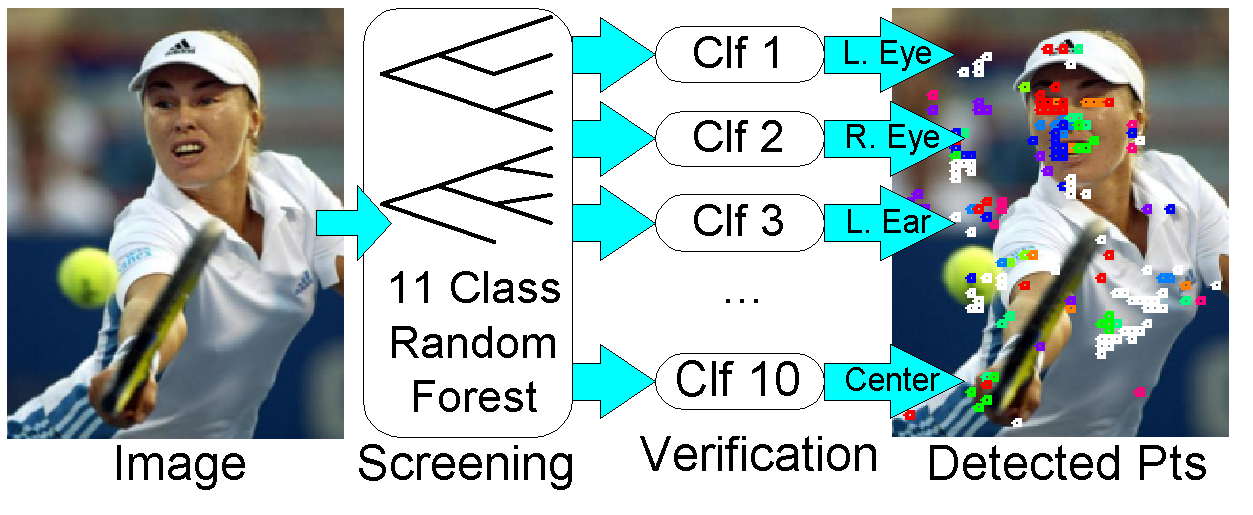}
\vskip -4mm
\caption{Keypoint Detection involves class prediction with a Random Forest and verification with binary classifiers (one class vs. background).}
\label{fig:keyptdet}
\vspace{-4mm}
\end{figure}

The face keypoints are detected in two stages as shown in Figure \ref{fig:keyptdet}. First, a 11-class (background, 9 keypoints, and face center) Random Forest with 100 trees of depth 11 prunes the search space. The trees use Aggregate Channel Features (ACF)\cite{dollar2014fast} with 10 channels and block size of 2 in a window of size $24\times 24$.

A FSA classifier \cite{barbu2013feature} with 3000 features selected from a pool of 61000 Haar and ACF features in a 24$\times$24 window is trained for each of the 9 keypoints and the face center, to minimize the Lorenz loss \eqref{eq:lorenz} and 10 iterations of hard negative mining \cite{felzenszwalb2010object}. These detectors have a detection rate on the training set of $90-95\%$, a false positive rate of $0.1-1\%$, and are used to verify the non-background classes proposed by the RF classifier. Other classifiers such as based on CNN could also be used.

\subsubsection{Generating 3D Pose Candidates}\label{sec:cg2}

Since the keypoints are detected for faces in a range of scales, the pose candidates are also obtained for faces in the same range.
The 3D pose candidates are generated by image-based regression from the detected keypoint locations. The 3D pose $\theta=(\bu,s,A)$ has six parameters $(\bu,s,\bphi)=(u^x,u^y,s,\varphi^x,\varphi^y,\varphi^z)$, where $\bphi=(\varphi^z,\varphi^x,\varphi^y)$ is the roll-pitch-yaw decomposition of the rotation matrix $A$. 

\noindent{\bf Image based regression.} The 3D pose is predicted from a point $(x_0,y_0,s_0)$ by image based regression, predicting the relative vector 
\vspace{-2mm}
\begin{equation}
(\frac{u^x}{s_0}\hspace{-0.7mm}-\hspace{-0.7mm}x_0,\frac{u^y}{s_0}\hspace{-0.7mm}-\hspace{-0.7mm}y_0,\frac{s}{s_0},\varphi^x\hspace{-1mm},\varphi^y\hspace{-1mm},\varphi^z)=\by(\bx).\label{eq:relpose}
\vspace{-2mm}
\end{equation}

The image based regression is trained using a feature vector $\bx=(x_1,...,x_m)$ selected from the same feature pool of $M=61,000$ features used in  Section \ref{sec:kpts} for face keypoint verification. The range of each variable is divided into $32$ equal bins and let $b_i(x)$ be the bin index function for variable $i$.
The regression vector $\by(\bx)$ is obtained as a sum of piecewise constant functions that depend on one variable each, i.e.
\vspace{-2mm}
\[
\by(\bx)=\sum_{i=1}^m \bz_{i,b_i(x_i)}
\vspace{-2mm}
\] 
where the $\bz_{ib}$ is the 6D coefficient vector for variable $i$ and bin $b$. All the coefficient vectors are collected into a 3 dimensional matrix $Z$ of size $6\times m \times 32$ that will be estimated from training examples.

\noindent{\bf Ground truth 3D pose.}  
The ground truth 3D poses are obtained by least squares energy minimization as described in section \ref{sec:3dgt}. The POSIT algorithm \cite{dementhon1995model} could also be used for this purpose.
Then the ground truth vectors for training the 3D pose regressors are obtained as in eq. \eqref{eq:relpose} for each annotated face from the fitted 3D pose 
$(\bu,s,\bphi)=(u^x,u^y,s,\varphi^x,\varphi^y,\varphi^z)$ and the keypoint location $(x_i,y_i,s_i)$. 

\noindent{\bf Training details.} The training examples are in the form of $(\bx_i,\by_i)\in \RR^M \times \RR^6$ where $\bx_i$ are feature vectors extracted at locations within 1 pixel from true keypoint locations and $\by_i$ are the relative vectors \eqref{eq:relpose} based on the ground truth poses obtained as described above. 
Training is done by minimizing the energy 
\vspace{-2mm}
\begin{equation}
L(Z)=\sum_{j=1}^n \|\by_j - \sum_{i=1}^m \bz_{i,b_i(x_{ji})}\|^2 \label{eq:lossreg}
\vspace{-2mm}
\end{equation}
using the FSA algorithm \cite{barbu2013feature} where $m=2000$ features are selected from the $M=61,000$ feature pool. A specific 6D pose regressor is trained for each keypoint for better accuracy. Observe that by using the loss function \eqref{eq:lossreg} the same 2000 features are used for predicting all six pose parameters, instead of using 2000 (possibly different) features for predicting each parameter. 

The percentage of variance explained $R^2$ for the pose regression dimensions ranges between 0.3 and 0.8, where the lowest scores were for predicting the pitch and roll angles. We choose FSA because it produced better 3D pose candidates than Random Forest regression, as will be seen in Sections \ref{sec:candeval} and \ref{sec:faceres}.

\subsubsection{Generating Face Candidates}\label{sec:facecand}

The 3D pose candidate generation step from the previous section obtains a number of 3D pose candidates $\theta_1, ... \theta_n$. The face candidates need to contain besides the 3D poses the predicted locations in 2D of the face keypoints.

From a 3D pose $\theta$ the keypoint locations $P=(\bp_1,...,\bp_L)$ are first predicted directly from the rigid 3D model as $P=T_{\theta}(R)$ , even though this prediction might not be very accurate. 
The IED (inter-eye distance) of this 3D candidate can also be computed from the 3D position of the eyes before projection.

\noindent{\bf Keypoint support.} The coherence term from eq. \eqref{eq:prior} is based on the number of keypoints that support the candidate. 
The support for candidate $(P,\theta)$ at some scale of the image pyramid is the number of rescaled $\bp_i$ that have a keypoint detection of type $i$ in the image at that scale at distance at most $0.1\cdot$IED (also rescaled to that scale). The overall support for the candidate is the maximum (over all scales) of the candidate support at one scale. The face keypoints that support the candidate become the new positions of the corresponding $\bp_i$.

The face candidates with a support less than a value $N^{supp}$ are eliminated. The value of $N^{supp}$ is usually 4 in our experiments, unless otherwise specified.
The number of generated face candidates for a 480$\times$320 image ranges ranges from about 4000 for $N^{supp}=1$ to about 1200 for $N^{supp}=4$.

\subsubsection{Scoring the Face Candidates} \label{sec:score}

A face candidate $F=(P,\theta)$ contains the 3D pose $\theta=(\bu,s,A)$ and predicted locations $P=(\bp_1,...,\bp_L)$ of the L keypoints. 
The score is obtained by a parameter-sensitive model depending on the yaw angle $\varphi^y$, as described in Section \ref{sec:parmodel}. 

\begin{figure}[ht]
\vspace{-4mm}
\centering
\includegraphics[height=1.3in]{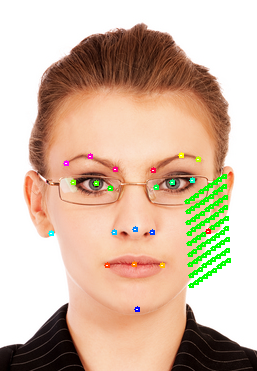}
\includegraphics[height=1.3in]{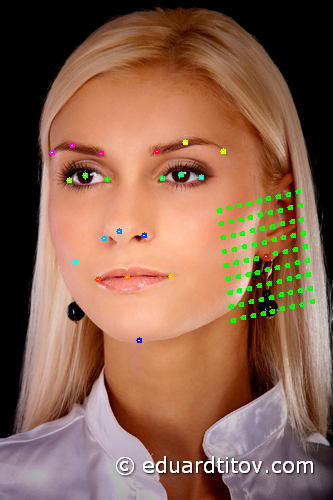}
\includegraphics[height=1.3in]{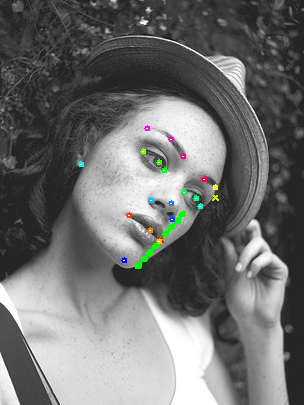}
\vskip -2mm
\caption{Sampling grid patterns based on the 3D face pose used to generate local features.}
\label{fig:grid}
\vspace{-3mm}
\end{figure}

\noindent {\bf Local difference features.} Inspired by the Local Binary Features (LBF) \cite{ren2014face} we generate a number of difference features relative to the 3D pose $\theta$ of a face candidate $F=(P,\theta)$ instead of the 2D shape. For that, an approximate tangent plane at each keypoint of the 3D face model is assigned a system of coordinates. This coordinate system is projected to a 2D coordinate system based on the 3D pose, which is used to define a skewed point grid centered at the predicted keypoint location $\bp_i$ as illustrated in Figure \ref{fig:grid}. 
On a $11\times 11$ grid we obtain about 72,000 such difference features around each keypoint.

\noindent {\bf Modified LBF features.} Modified LBF features can be obtained as the leaf indexes for the Random Forest regression trees trained with these local difference features. The LBF were trained with 100 Random Trees of depth 6 for each of the 9 keypoints, for a total of $100\times 32\times 9=28,800$ features.

\noindent {\bf Local selected features (LSF).} Another approach we will use is to first prune the 72,000 local difference features at each keypoint to a more manageable size (e.g. 1,000) which will be the input of the parameter sensitive model described in Section \ref{sec:parmodel}. The pruning of the local difference features is done by FSA regression to predict local translations of the grid center on the training examples, similar with the LBF training but using FSA instead of RF for regression.

\noindent {\bf Special features.} For each face candidate we also created 2 special features for each keypoint including the face center. First is the distance (as percentage of the inter-eye distance of the candidate) between the predicted keypoint location of the candidate and the closest detected keypoint to that location. If there is not such detected keypoint or the distance is larger that $0.32$ IED, the value is $0.32$.  The second feature is the detection score of the closest detected keypoint, with a minimum of $-1.1$. For the 9 keypoints plus face center we obtain this way 20 special features. We also added the candidate support as the 21-st special feature.

In both cases, let $\bx(P,\theta)$ be the vector of LBF or LSF features+special features extracted from the image for the candidate $F=(P,\theta)$.

\noindent {\bf Score function.} The score $S(P,\theta)$ of the 3D face candidate with face keypoints $P$ and pose $\theta$ is based on the feature vector $\bx(P,\theta)$, either by a linear model \eqref{eq:scorelbf} for the LBF features or by a nonlinear model \eqref{eq:scorelsf} for the LSF features.

The models depend parametrically on the yaw angle $\varphi^y$ of  the rotation $A$. The yaw angle ranges between $-\pi/2$ and $\pi/2$, being $0$ for frontal faces and $\pm\pi/2$ for profile faces. For this application, it is discretized into $B_\theta=16$ bins for the LBF features and $B_\theta=9$ bins for the LSF features.

\subsubsection{Non-Maximal Suppression}

The non-maximal suppression step iterates the following until convergence:

1. Select the face candidate $(P,\theta)$ with the largest score above a threshold and finds the bounding box $B$ of its points $P$.

2. Remove the candidates that have at least $50\%$ overlap with $B$.

\section{Fitting 3D Models}\label{sec:prereq}

In this section we show how the 3D model is fitted to and learned from 2D annotations.
\vspace{-2mm}
\subsection{Fitting a Rigid Projection Transformation.}
Given a matrix $3\times L$ matrix $F$ and a set of 2D points $P=(\bp_1,...,\bp_L)$ in the form of a $2\times L$ matrix, the goal is to find a rigid transformation $\theta=(\bu,s,R)$ to minimize:
\vspace{-2mm}
\[
E(\bu,s,R)=\|\bu\bo+s\pi(RF)-P\|^2
\vspace{-2mm}
\]
where $\pi((\bx,\by,\bz)^T)=(\bx,\by)^T$ and $\bo\in \RR^L$ is the row vector with all entries $1$.

The algorithm uses hidden variables for the $z$ coordinates of the points $\bp_i$ and iterates fitting the rigid transformation with updating the $z$-values.
\vspace{-3mm}
\begin{algorithm}[htb]
   \caption{{\bf Fit Rigid Projection}}
   \label{alg:rigidproj}
\begin{algorithmic}
   \STATE {\bfseries Input:} $3\times L$ matrix $F$ and $2\times L$ matrix $P$.
   \STATE {\bfseries Output:} Scalar $s$, $3\times 3$ rotation matrix $R$ and 2D vector $\bu$ to minimize $\|\bu\bo+s\pi(RF)-P\|^2$
\end{algorithmic}
\begin{algorithmic} [1]
\STATE Initialize $L\times 3$ matrix $B=(P^T,0)$.
\FOR {i=1 to $N^{iter}$} 
\STATE Call Algorithm \ref{alg:rigid} to find $\bu, s,R$ to minimize $\|\bo^T\bu^T+sF^TR-B\|^2$
\STATE Extract third column $\bc_3=(C_{i3})_{i}$ of $C=sF^TR$
\STATE Update $B=( P^T,\bc_3)$
\ENDFOR
\STATE Change $R$ to $R^T$ and discard the $z$-component of $\bu$.
\end{algorithmic}
\end{algorithm}
\vspace{-3mm}

The algorithm to fit a rigid transformation between two sets of points of the same dimension $d$ is due to Schonemann \cite{schonemann1970fitting} and is presented in Algorithm \ref{alg:rigid}.
\vspace{-3mm}\begin{algorithm}[htb]
   \caption{{\bf Fit 3D Rigid Transformation}}
   \label{alg:rigid}
\begin{algorithmic}
   \STATE {\bfseries Input:} Matrices $A,B$ of size $p\times d$.
   \STATE {\bfseries Output:} Scalar $s$, $d\times d$ rotation matrix $R$ and $d\times 1$ vector $\bu$ to minimize $\|\bo^T\bu^T+sAR-B\|^2$
\end{algorithmic}
\begin{algorithmic} [1]
\STATE Compute the column means $\bar \alpha= \bo A/p,\; \bar \beta=\bo B/p$ and column centered matrices $A^*=A-\bo^T \bar \alpha$ and $B^*=B-\bo^T\bar\beta $.
\STATE Decompose $A^{*T}B^*=UDV^T$ by SVD, where $U,V$ are rotation matrices and $D$ is a diagonal matrix.
\STATE Obtain $R=UV^T$, $\bu=\bar \beta-s\bar\alpha R$ and
\vspace{-3mm}
\[
s=tr[R^TA^{*T}B^*]/tr(A^{*T}A^*)
\vspace{-3mm}
\]
\end{algorithmic}
\end{algorithm}
\vspace{-3mm}

 Given  two sets of points $A,B$ of the same dimension $d$, it finds a rigid transformation $(\bu, s,R)$ represented by a translation vector $\bu$, scaling $s$, and rotation matrix $R$ and  to minimize $\|\bo^T\bu^T+sAR-B\|^2$.

\begin{table*}[htb]
\vspace{-7mm}
\small
\begin{center}
\caption{Evaluation of face candidates on the FDDB dataset.}\label{tab:detcand}
\begin{tabular}{|c|c|c|c|c|c|c|c|c|c|}
\hline
Experiment &Method & &Keypoint &Pose Regression&$\phantom{I^I}$ & \multicolumn{2}{|c|}{False Positive Rate} & \multicolumn{2}{|c|}{Detection Rate}\\
Number &Name &Keypoints &Verification &Method  & $N^{supp}$ &\%$<$0.3 &\%$<$0.5 &\%$>$0.5 &\%$>$0.7\\
\hline
1 & S0S0 &Center only &FSA &FSA &0 &60.2 &77.5 &96.6 &90.8\\
2 &R9S3 &9+Center &- &FSA &3 &47.7 &70.6 &90.0 &86.4\\
3 &RS9S2 &9+Center &FSA &FSA &2 &28.8 &52.1 &97.1 &95.0\\
4 &RS9S3 &9+Center &FSA &FSA &3 &9.2 &30.4 &94.0 &91.7\\
5 &RS9S4 &9+Center &FSA &FSA &4 &2.3 &16.6 &90.5 &88.1\\
6 &RS9R3 &9+Center &FSA &RF   &3 &12.1 &35.5 &94.0 &90.8\\
\hline
\end{tabular}
\end{center}
\vspace{-5mm}
\end{table*}

\subsection{Learning a 3D Model from 2D Annotations} \label{sec:3dgt}

The face 3D model matrix $R$ is obtained from  a number of 2D face images where the $L$ keypoints have been annotated.

Let $P_i=(p_{i1},...,p_{iL})$ be the 2D coordinates of the $L$ keypoints for face $i, i=\overline{1,n}$. Write $P_i=\begin{pmatrix}\bx_i \\ \by_i\end{pmatrix}$, obtaining the row vectors $\bx_i,\by_i$ as the  $x$ and $y$ coordinates of the $L$ keypoints.

The goal is to find the matrix $F$ of size 3$\times$L and the projected rigid transformations $\theta_i, i=\overline{1,n}$ for the annotated faces, to minimize
\vspace{-3mm}
\begin{equation}
\begin{split}
E(\theta,F)=\sum_{i=1}^n \|\bu_i\bo+s_i\pi(R_iF)-P_i\|^2\\
=\sum_{i=1}^n \|T_i^x F - X_{i}\|^2+\sum_{i=1}^n \|T_i^y F - Y_{i}\|^2 \label{eq:fitfaces}
\end{split}
\end{equation}
where each $T_{\theta_i}$ is a $2\times 3$ matrix with rows $T_i^x, T_i^y$.

The minimization starts with a random F and alternates two steps until convergence:
\begin{enumerate}
\item Given current $F$, fit the projected rigid transformations $\theta_i$ using Algorithm \ref{alg:rigidproj} for each face
\item Given current $\theta_i$, find F by minimizing eq. \eqref{eq:fitfaces}
\end{enumerate}

This approach is a simplified version of the Stratified Procrustes Analysis \cite{bartoli2013stratified}.

\vspace{-2mm}
\section{Experiments}

{\bf Method nomenclature.} The proposed method consists of three main parts: a) face keypoint and center detection, b) 3D pose regression+ support pruning and c) face candidate evaluation. Since each has a number of variants, we will represent each combination with a shortcut:
\vspace{-2mm}\[
 \underbrace{RS9}_{a)} \underbrace{S4}_{b)} \underbrace{PLSF}_{c)}
 \vspace{-4mm}\]
where
\begin{itemize}
\item [a)] Keypoint detection can be R (Random Forest), S (FSA) or RS (Random Forest +FSA), and 9 is the number of keypoints that will be detected (plus face center). The RF screening and the FSA verification are covered in section \ref{sec:kpts}.
\item [b)]  3D Pose regression can be R (Random Forest) or S (FSA), as described in Section \ref{sec:cg2}. The number $4$ comes from pruning the 3D candidates by the support, as described in Section \ref{sec:facecand}, thus $N^{supp}=4$.
\item [c)]  Face candidate scoring (verification) has been described in Section \ref{sec:score}, and it can be PLSF = Parametric sensitive model with LSF+special features, PLBF = parametric sensitive model with LBF features, or LBF= standard LBF classifier.
\end{itemize}

{\bf Training dataset.} For training we used 11,000 images from the AFLW dataset \cite{koestinger11b}, containing about 14,300 faces. All models (keypoint detectors, 3D model, LBF features, parameter sensitive classifier, etc) were trained on these images and their 21 point annotation.

\vspace{-4mm}
\subsection{Evaluation of Face Candidates}\label{sec:candeval}
\vspace{-1mm}

Before evaluating the whole system, we first evaluate the face candidate generator to get an idea of what to expect from the face verification step.
For a face candidate $F=(P,\theta)$ with 3D pose $\theta$ one can compute a face bounding box based on the predicted keypoints $P$.
The face candidates can be evaluated by computing the overlap of the face bounding boxes with the ground truth face bounding boxes. 
\begin{figure*}[t]
\centering
\hspace{-3mm}
\includegraphics[height=5.5cm, width=7.5cm]{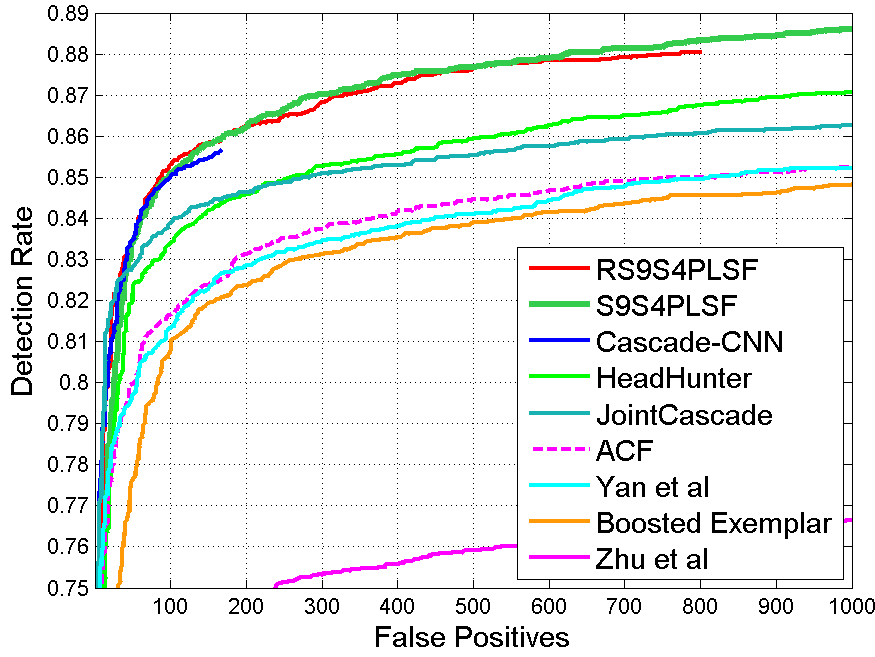}
\hspace{-2mm}
\includegraphics[height=5.5cm, width=7.5cm]{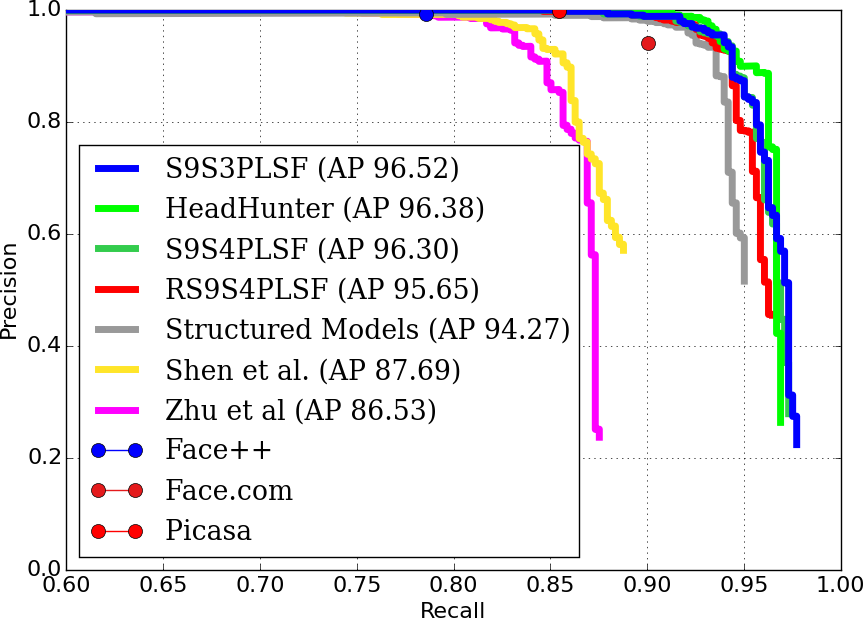}
\vskip -3.mm
\caption{Results and comparisons with other methods. Left: the FDDB dataset ( 2845 images with 5171 faces). Right: the AFW dataset (205 images with 552 faces).}
\label{fig:fddb}
\vspace{-7mm}
\end{figure*}

In Table \ref{tab:detcand} are shown two measures of ``false positive rate'': the percentage of candidate boxes with overlap $<0.3$ or $<0.5$ with the ground truth face bounding boxes and two "detection rate" measures: the percentage of GT faces that have candidate bounding boxes with overlap $>0.5$ or $>0.7$.

To see the importance of the keypoint detections, experiment 1 shows the face candidates obtained only from the face center and no support pruning.
Experiment 4 has the candidates predicted from all nine keypoints plus the center and $N^{supp}=3$, clearly better than the ones from the center only in both higher detection rate  (91.7\% vs 90.8\%) and lower false positive rate (9.2\% vs 60.2\%). Experiment 2 shows the candidates obtained without the keypoint verification step from Figure \ref{fig:keyptdet} are clearly inferior to Experiment 4.
Experiment 6 shows the pose regression based on a Random Forest with 100 trees of depth 10 using the same features as the FSA pose regression. One can see that the FSA pose regression from Experiment 4 obtains higher detection rate  (91.7\% vs 90.8\%) and lower false positive rate (9.2\% vs 12.1\%).

\vspace{-3mm}
\subsection{Face Detection Results} \label{sec:faceres}

We present results on two standard face detection datasets: The FDDB dataset \cite{fddbTech} with 2845 images containing 5171 faces and the AFW dataset \cite{zhu2012face} with 205 images and 552 faces. The FDDB evaluation used the evaluation code provided on the FDDB website. The AFW evaluation used the code provided by \cite{mathias2014face}. In both cases a detection with less than 50\% overlap with any annotated face is considered a false positive.

\begin{figure*}[ht]
\vspace{-9mm}
\centering
\includegraphics[height=4.cm]{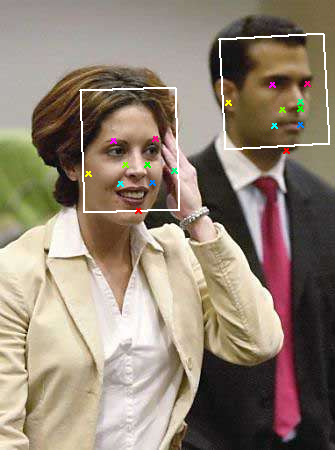}
\includegraphics[height=4.cm]{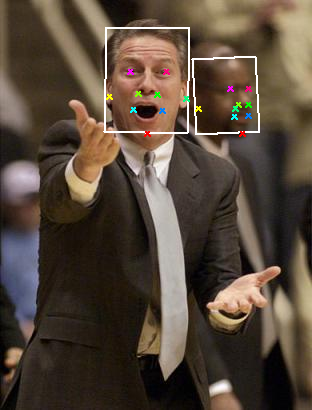}
\includegraphics[height=4.cm]{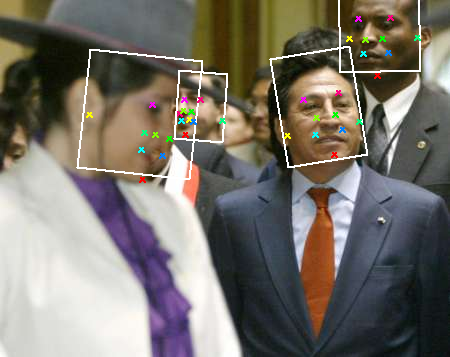}
\includegraphics[height=4.cm]{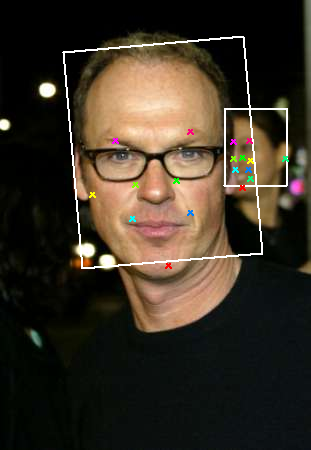}
\includegraphics[height=4.cm]{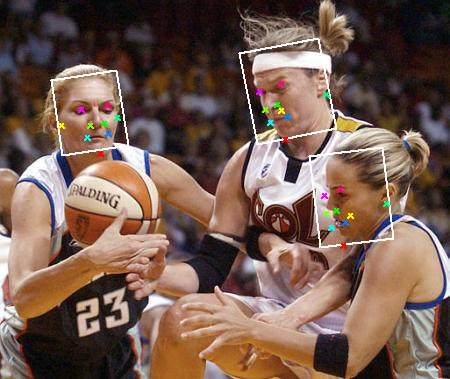}
\includegraphics[height=4.cm]{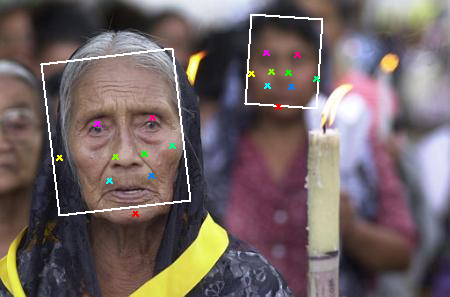}
\includegraphics[height=4.cm]{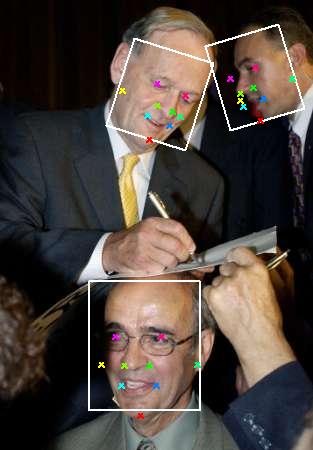}
\includegraphics[height=4.cm]{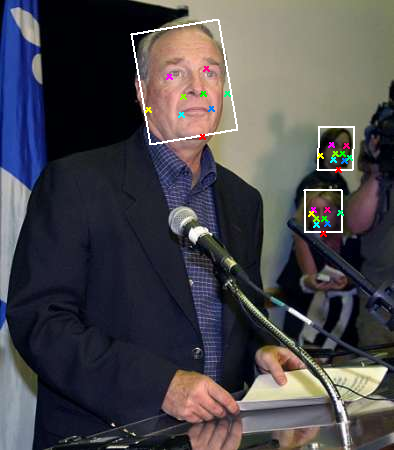}
\includegraphics[height=4.cm]{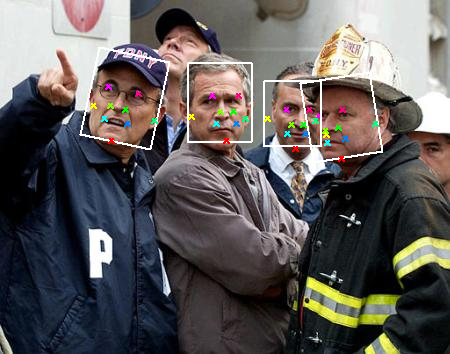}
\includegraphics[height=4.cm]{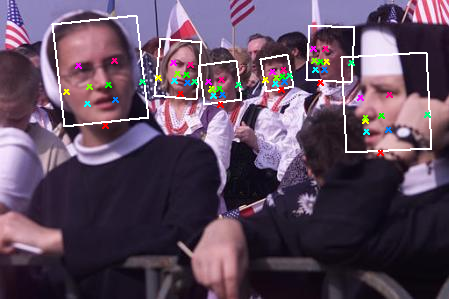}
\includegraphics[height=4.cm]{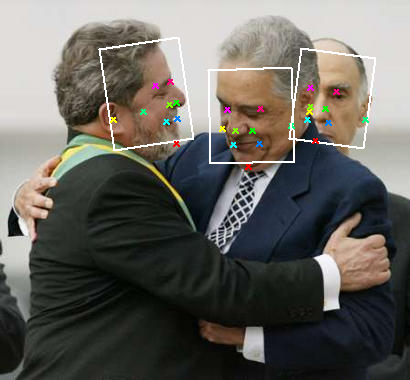}
\includegraphics[height=4.cm]{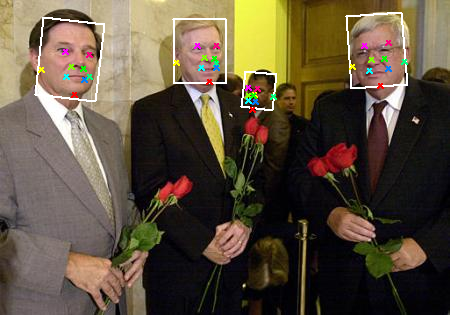}
\includegraphics[height=4.cm]{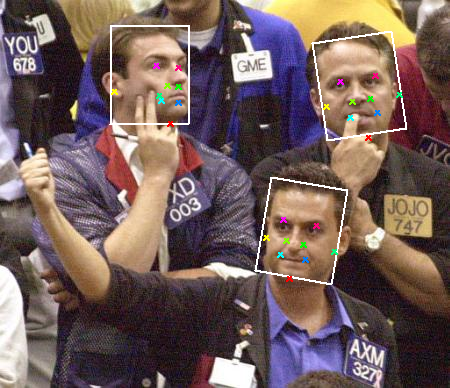}
\includegraphics[height=4.cm]{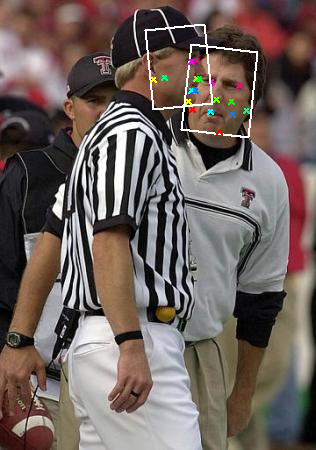}
\includegraphics[height=4.cm]{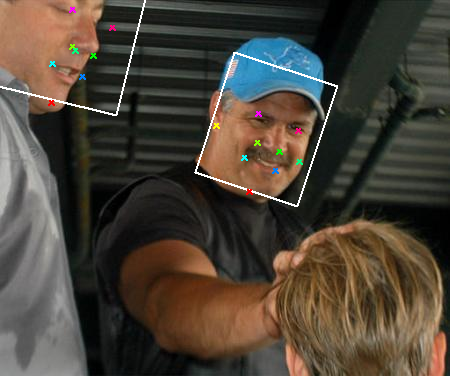}
\includegraphics[height=4.cm]{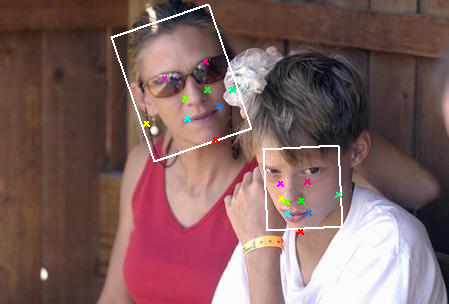}

\vskip -1mm
\caption{Detected faces on the FDDB dataset.}
\label{fig:results}
\vspace{-3mm}
\end{figure*}

The results on the FDDB dataset are shown in Figure \ref{fig:fddb}, left. Our results are: ``S9S4PLSF''  and ``RS9S4PLSF'' with LSF+special features with FSA respectively RF+FSA keypoint detection, FSA pose regression and $N^{supp}=4$. Also shown are results from the Cascade-CNN \cite{li2015convolutional}, Joint Cascade \cite{chen2014joint}, HeadHunter \cite{mathias2014face}, Boosted Exemplar \cite{li2014efficient}, ACF \cite{yang2014aggregate}, Yan et al \cite{yan2014fastest} and Zhu \cite{zhu2012face}.
One can see that the proposed method obtains results comparable to the state of the art for up to 100 false positives and outperforms the other methods for the regime with at least 100 false positives. In Figures \ref{fig:results} and \ref{fig:results2} are shown detection results on a few images of the FDDB dataset. 

The results on the AFW dataset are shown in Figure \ref{fig:fddb}, right. Also shown are results from the Head Hunter \cite{mathias2014face}, Shen et al \cite{shen2013detecting}, Structured Models \cite{yan2014face} and Zhu \cite{zhu2012face}. The algorithm performs well in the high recall regime and is outperformed by the Head Hunter \cite{mathias2014face} in the high precision regime.
\begin{figure}[htb]
\hspace{-1mm}
\includegraphics[width=7.5cm]{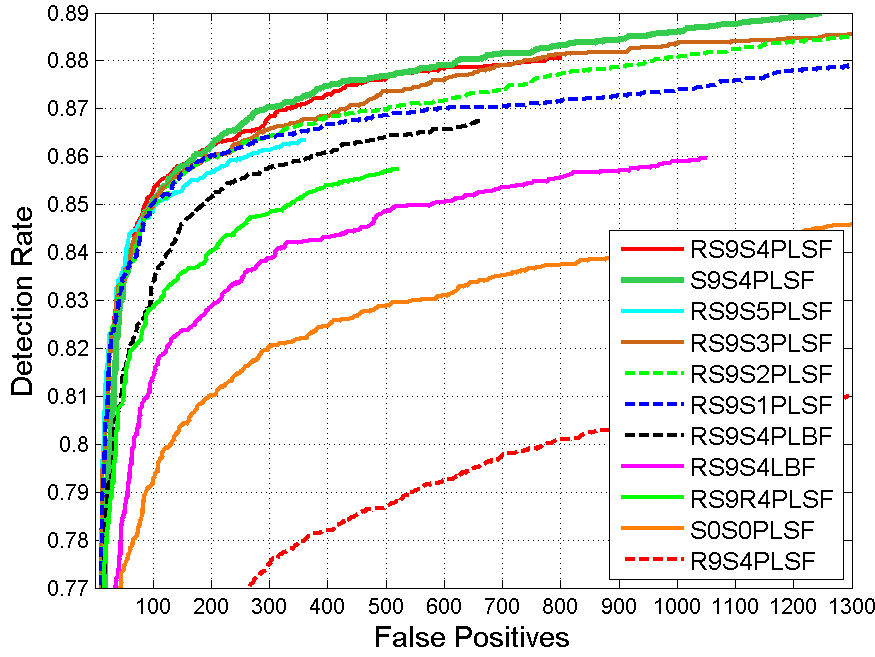}
\vskip -3mm
\caption{Evaluation of design decisions on FDDB.}
\label{fig:design}
\vspace{-3mm}
\end{figure}

\noindent{\bf Evaluation of design decisions.} In Figure \ref{fig:design}, are shown evaluations to support the decisions to use multiple keypoints,  FSA pose regression, support pruning, and a parameter sensitive classifier. 
The result ``S0S0PLSF'' has candidates predicted only from the detected  face centers, with $N^{supp} =0$. It shows that predicting poses from multiple keypoints and support pruning has a considerable increase in detection accuracy. The result ``R9S4PLSF'' detects keypoints with the  RF and without the FSA verification classifiers from Figure \ref{fig:keyptdet}, clearly showing the importance of the keypoint verification step.
The result ``RS9S4LBF'' uses a parameter insensitive classifier trained with the logistic loss, and performs inferior to the parameter sensitive classifier ``RS9S4PLBF'', which in turn is inferior to the parameter sensitive classifier ``RS9S4PLSF'' based on LSF+special features.
The result ``RS9R4PLSF'' uses the Random Forest described in Section \ref{sec:candeval} for 3D pose regression and the parameter sensitive classifier for verification. Again, it performs inferior to the algorithm ``RS9S4PLSF'' with FSA-based pose regression. Also shown are results with the proposed method with $N^{supp}=1,2,3,5$.
\begin{figure}[htb]
\vspace{-3mm}\centering
\includegraphics[width=7cm]{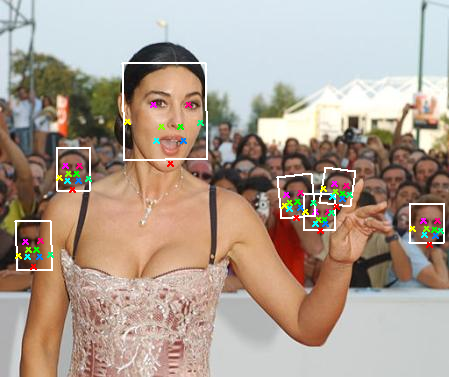}
\vskip -3mm
\caption{Failure example from the FDDB dataset where many small, blurry and occluded faces have not been detected.}
\label{fig:failures}
\vspace{-3mm}
\end{figure}

\noindent{\bf Failure modes.} We observed that the proposed 3D model-based face detection approach has difficulties mainly with detecting small, blurry and occluded faces, as shown in Figure \ref{fig:failures}. Indeed, a 3D model based on face keypoints doesn't make much sense for small faces since the keypoints might be too small to be visible.

\noindent{\bf  Detection time.} The detection time for a 480x320 image is about 3 seconds when using the RF screening and 15 seconds without the RF screening. The C++ code has not been optimized for speed and most of the time is used for detecting the keypoints. We expect to obtain speedups of 10-100 times with a GPU implementation and code optimization.

\vspace{-2mm}
\section{Conclusion}

In this paper we presented a method for face detection that uses a 3D model to represent the face hypotheses, extract 3D pose-aligned features and to specify the yaw parameter value for a parameter sensitive classifier. The 3D face candidates are proposed by image based regression starting from a number of face keypoints that are detected first.
From experiments we observed that generating 3D face candidates from multiple face keypoints and pruning them based on the number of keypoints detected results in considerable improvements in detection accuracy compared to generating candidates only from the face center. 
The results obtained are comparable with the state of the art in the low false positive regime and outperforms the cascade-based state of the art methods for the regime of at least 0.04 false positives per image.

\vspace{-1mm}
{\small
\bibliographystyle{ieee}
\bibliography{featselectbib}
}

\end{document}